\documentclass[10pt,twocolumn,letterpaper]{article}
\usepackage{cvpr}
\usepackage{times}
\usepackage{epsfig}
\usepackage{graphicx}
\usepackage{amsmath}
\usepackage{amssymb}
\usepackage{algorithm}
\usepackage{algorithmic}

\usepackage{booktabs}
\usepackage{color}
\usepackage{overpic}
\usepackage{contour}

\definecolor{tonacha}{RGB}{79,114,108}
\definecolor{yaqing}{RGB}{69,76,79}

\usepackage[pagebackref=true,breaklinks=true,letterpaper=true,colorlinks,bookmarks=false]{hyperref}

\cvprfinalcopy 

\def\cvprPaperID{1449} 

\ifcvprfinal\fi
\begin{document}

\title{Multi-Miner: Object-Adaptive Region Mining for Weakly-Supervised Semantic Segmentation}

\author{%
  Kuangqi Zhou$^1$, 
  Qibin Hou$^{1}$,
  Zun Li$^2$,
  Jiashi Feng$^1$,
  \\
  $^1$National University of Singapore, 
  $^2$Beijing Jiaotong University\\
  \texttt{kzhou@u.nus.edu}, 
  \texttt{\{andrewhoux,lznus2018\}@gmail.com},\\
  \texttt{elefjia@nus.edu.sg}\\
}

\maketitle

\begin{abstract}
Object region mining is a critical step for weakly-supervised semantic segmentation. Most recent methods mine the object regions by expanding the seed regions localized by class activation maps. They generally do not consider the sizes of objects and apply a monotonous procedure to mining all the object regions. Thus their mined regions are often insufficient in number and scale for large objects, and on the other hand easily contaminated by surrounding backgrounds for small objects. In this paper, we propose a novel multi-miner framework to perform a region mining process that adapts to diverse object sizes and is thus able to mine more integral and finer object regions. Specifically, our multi-miner leverages a parallel modulator to check whether there are remaining object regions for each single object, and guide a category-aware generator to mine the regions of each object independently. In this way, the multi-miner adaptively takes more steps for large objects and fewer steps for small objects. Experiment results demonstrate that the multi-miner offers better region mining results and helps achieve better segmentation performance than state-of-the-art weakly-supervised semantic segmentation methods. 
\end{abstract}

\section{Introduction}

\newcommand{\addTexB}[1]{\contour{white}{\textcolor{black}{#1}}}
\begin{figure}
    \centering 
    \begin{overpic}[width=0.48\textwidth]{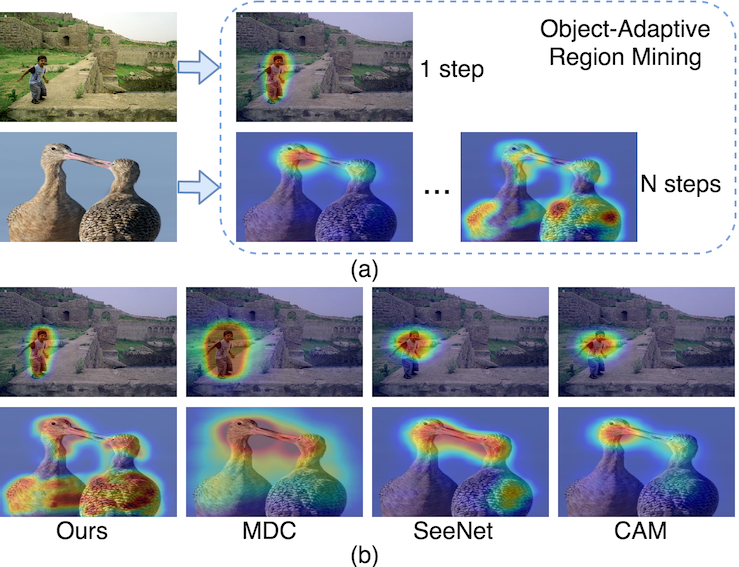}
    \put(0.5, 56){\addTexB{\scriptsize \sffamily bird}}
    \put(0.5, 73){\addTexB{\scriptsize \sffamily person}}
    \end{overpic}

\caption{(a) Object-adaptive region mining performed by our multi-miner. Our method adaptively takes more steps for large objects and fewer steps for small objects. (b) Comparisons among regions mined by our method and previous methods~\cite{mdc,seenet}, and the original seed regions localized by CAM~\cite{cam}. Our mined regions are more integral and finer. Better viewed in color.} 
\label{fig:intro-a} 
\end{figure}

Weakly-supervised semantic segmentation with image-level supervision is widely studied to relieve the scarcity of pixel-level annotations. Object region mining is the key step for recent weakly-supervised semantic segmentation methods~\cite{ae,seenet,mdc, gain,ficklenet}, which aims to expand the sparse seed object regions localized by class activation maps~\cite{cam,gradcam}.

In the existing {weakly-supervised semantic segmentation} methods, the regions of all the objects are mined in a monotonous manner, with a pre-fixed number of erasing steps~\cite{ae,seenet, gain} or randomly selected hidden units~\cite{ficklenet}, \etc. However, as the size of object regions varies for different objects in different images, the optimal number of region mining steps also differs. It can be observed that the mined regions of existing methods are often insufficient in number and scale for large objects and for small ones tend to be contaminated by surrounding backgrounds.

In this paper, we propose a novel multi-miner framework that can perform a region mining process fully adaptive to every single object. As shown in Fig.~\ref{fig:intro-a}, our method adaptively takes different region mining steps for different objects, and thus offers more integral and finer region mining results than existing methods. 

The key to such object-level adaptability is to automatically stop mining for an object when all of its regions are mined, and continue mining for an object whose regions are not completely mined. To achieve this, we leverage a parallel modulator, which can be simply implemented as a {multi-label} classifier, to control the region mining process by checking whether there are remaining object regions for each single object in an input image in parallel. The modulator then guides a category-aware generator to continue or stop mining regions for each object. With such a checking-mining mechanism, the number of region mining steps for each object adapts to its size. {Intuitively}, our multi-miner consists of multiple parallel ``region miners'', each of which performs the region mining sub-process for a single object. Furthermore, we also provide the theoretical basis of the multi-miner by revealing the connection between region mining and distribution mapping.

\begin{figure}
\centering 
\includegraphics[width=0.48\textwidth]{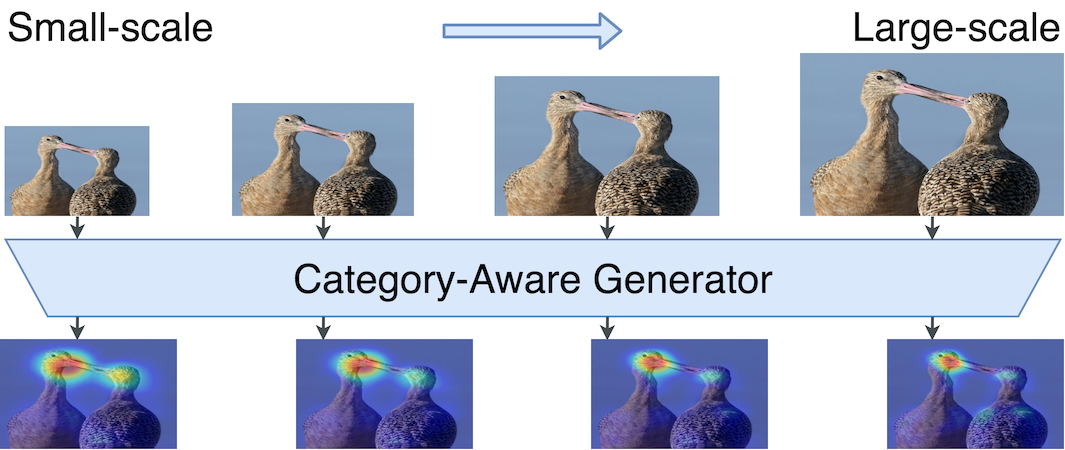}
\caption{Regions mined in one step with inputs of different scales. Better viewed in color.} 
\label{fig:intro-b} 
\end{figure}

Some backgrounds would be included inevitably during region mining as the spatial resolution of input images is not preserved due to stacked convolutional layers. To alleviate this problem, we further propose a multi-scale training strategy to help our multi-miner progressively mine finer regions. Our motivation is, small-scale inputs provide global information about the whole objects, while large-scale inputs provide information about details. As illustrated in Fig.~\ref{fig:intro-b}, as the scale of the input image increases, the category-aware generator mines smaller but finer object regions. Thus, the multi-miner roughly mines the objects during early steps, and then further mines the small remaining regions in later steps. The regions mined in later steps are smaller and finer, and include fewer background regions.

We apply the multi-miner to mining object regions from input images and train a semantic segmentation model with our mined regions. Extensive experiments demonstrate that the multi-miner provides higher-quality region mining results and offers better segmentation performance, compared with well-established baselines. In particular, it helps achieve the mIoU of 65.9\% and 66.1\% on PASCAL VOC 2012 \textit{val} and \textit{test} sets.

To sum up, we make the following contributions:
 \begin{itemize}
 \setlength\itemsep{0em}
 
  \item We propose a multi-miner, which is the first model to perform region mining procedure that is fully adaptive to different objects. We also provide the theoretical basis for our multi-miner.
  
 \item We propose a multi-scale training strategy to work with the multi-miner to mine progressively finer object regions and prohibit backgrounds from being mined.
 
 \item Experiment results on PASCAL VOC 2012 segmentation benchmark show that our method establishes new state-of-the-art  under the same weakly-supervised setting. 

 \end{itemize}
 
\begin{figure*}[th]
\centering 
\includegraphics[width=1\textwidth]{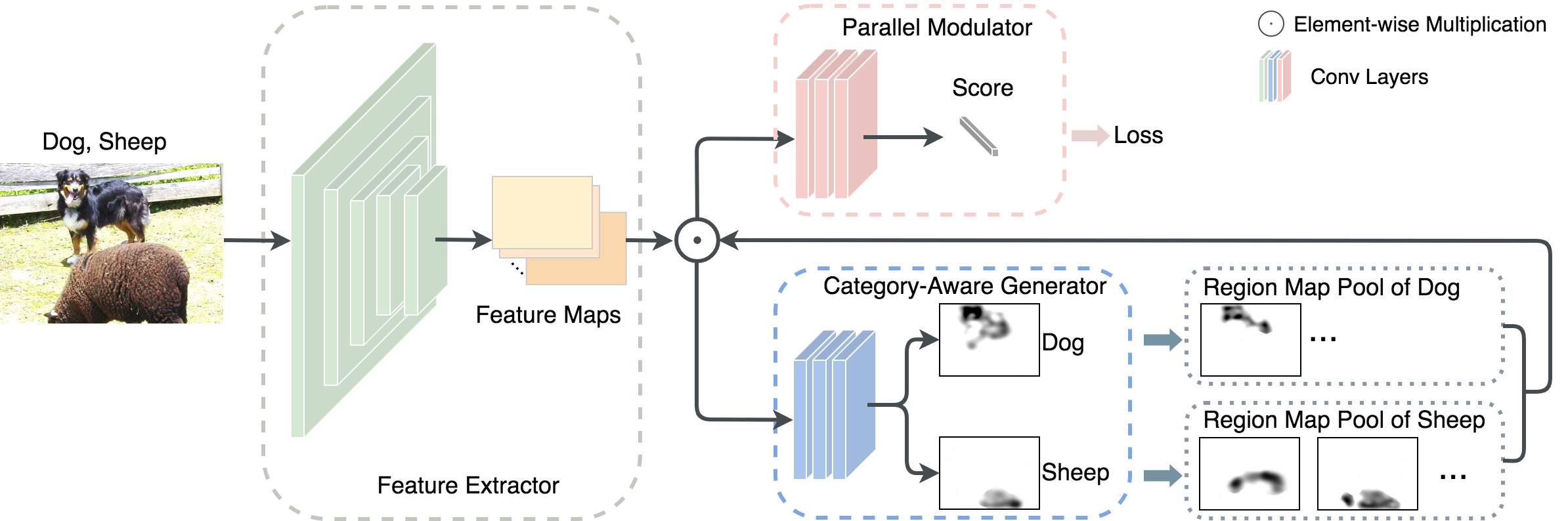}
\caption{Model architecture of the multi-miner. The category-aware generator generates region maps to mine the object regions. The parallel modulator {controls} the whole region mining procedure by performing a multi-label classification task. Both the category-aware generator and the parallel modulator are trained via the loss of the modulator. A region map pool is maintained for each object to store all region maps.} 
\label{fig:model} 
\end{figure*}

\section{Related Work}
A variety of weakly-supervised methods have been proposed to mitigate the insufficiency of pixel-level supervision in semantic segmentation. For example, Dai \etal~\cite{dai_bbx} and Papandreou \etal~\cite{ppdr} propose to leverage bounding boxes as supervision for semantic segmentation; Lin \etal~\cite{scribbles} take scribbles as relatively coarse labels.  Compared with bounding boxes and scribbles, image-level labels are easier to obtain and thus widely exploited. In this work we focus on weakly-supervised semantic segmentation with image-level supervision.

Some early methods propose to directly train a segmentation model with image-level labels~\cite{ppdr, pthk1, pop, wildcat} through Multiple Instance Learning (MIL)~\cite{mil}. However, as image-level labels do not include sufficient information for learning a segmentation model, the performance of these methods is not satisfactory. 
Recently, most methods~\cite{sec, ae, dcsp, seenet, ficklenet} propose to mine object regions by expanding the seed regions localized by the class activation maps~\cite{cam, gradcam}. These methods can be divided into the following two categories. 

The first category, like \cite{sec, dsrg, affinitynet}, expand the object regions outside the training process of the classification model. SEC~\cite{sec} proposes to refine the seed regions by an approximated CRF~\cite{crf}. In~\cite{dsrg}, seed regions are refined in an unsupervised manner during the segmentation model training. AffinityNet~\cite{affinitynet} leverages the seed regions to train an additional network for learning pixel-level affinity, which is in turn used to expand the seed regions.

The second category, such as \cite{ae, seenet, gain, mdc, mcof, ficklenet}, mine object regions during the training process of the classification model. AE~\cite{ae} expands the seed regions by iteratively erasing the newly found discriminative regions and then re-training the classification model. \cite{gain} modifies the erasing strategy to be end-to-end trainable. Furthermore, SeeNet~\cite{seenet} introduces two self-erasing strategies to keep unexpected background regions from being discovered. \cite{mdc} revisits the dilated convolution \cite{deeplabv1} and mines object regions by merging the feature maps generated by multiple dilated convolutional blocks of different dilation rates. In~\cite{mcof}, object regions are mined by iteratively training the classification model and the segmentation model. More recently, FickleNet~\cite{ficklenet} proposes to discover different object regions by randomly selecting various hidden units to generate attention maps. 

All these methods mine the object regions for all the objects in a monotonous way. Different from them, our method adaptively takes different numbers of region mining steps for different objects.

\section{Proposed Method}
\subsection{Object-adaptive region mining}
\paragraph{Architecture of {the multi-miner}}
The architecture of the proposed multi-miner is illustrated in Fig.~\ref{fig:model}. It consists of three components: a backbone feature extractor, a parallel modulator and a category-aware generator. 
The feature extractor is used to extract high-level features of the input images.
The parallel modulator aims to control the region mining procedure by checking whether there are remaining object regions left for each category of each image, which can be exactly implemented by a multi-label classification model. The multi-label classifier is actually a combination of multiple independent binary classifiers, each of which corresponds to one semantic category. For each semantic category, if there are still remaining object regions, the classifier will recognize the object continuously based on the regions. Otherwise, the classifier will not be able to recognize this object.
This characteristic can be leveraged to guide the category-aware generator to continue or stop mining object regions. 
The category-aware generator is fully-convolutional. The number of channels of its last layer is equal to the number of all possible categories, such that it can generate a region map for each single category in the image. A region map is used to mine the object regions in the input image via element-wise multiplication; the pixel values of a region map are in the range $[0, 1]$. In a region map, the mined regions have low pixel values.

\paragraph{Region mining procedure}
The multi-miner adaptively takes different numbers of region mining steps for different objects. In each step, the multi-miner mines object regions via the following three stages. First, the parallel modulator is updated to evaluate the remaining object regions for all semantic categories in the input image. Second, under the guidance of the updated modulator, the category-aware generator is trained to continue or stop mining regions for each semantic category.  Third, the updated category-aware generator is used to generate region maps. 
To reduce computation cost, we apply the region maps to the output feature maps of the feature extractor, instead of the input images.

Formally, let $\mathcal{I} = \{ (I_i, y_i)\}_{i=1}^{N} $ denote the training set of $N$ images, where $y_i$ is the image-level label for  image $I_i$\footnote{For simplicity, we omit the subscript $i$ unless necessary in the following part of this paper}.  Denote the set of all possible object categories as $\mathcal{C}$, and the set of image-level labels in a given image $I$ as $\mathcal{C}_{pos}$. We use $\theta_e$, $\theta_m$ and $\theta_g$ to represent the parameters of the feature extractor, parallel modulator, and category-aware generator respectively. Additionally, we maintain a region map pool $P_{j}$ for each semantic category $j\in \mathcal{C}_{pos}$ in each image $I$ to store the region maps in all steps.

In the first stage of the $t$-th region mining step, the feature extractor takes in $I$ and outputs the feature map $F$. We mask the previously mined regions from $F$, obtaining $F^t$:
\begin{equation}
  F^t = F \odot \min\limits_{\tau, j}\{M^\tau_{j}\}.
  \label{eqn:feature masking}
\end{equation}

Here $\odot$ denotes an element-wise multiplication, and $M^\tau_j$ denotes the region map of semantic category $j$ generated in the $\tau$-th iteration $(\tau=1,\cdots, t-1)$. The minimization operation is always assumed to be conducted individually for every single spatial location unless stated otherwise. In Eqn.~\eqref{eqn:feature masking}, we adopt the minimization operation at every spatial location across all previously generated category-specific region maps for two reasons. First, object regions of one category are regarded as backgrounds for other categories. {Therefore, such a Winner-Take-All merging strategy avoids object regions of one category from being overridden by region maps of other categories. }This ensures that our multi-miner mines regions for all semantic categories independently. Second, this strategy prohibits interference of the regions mined in different steps.

We train the parallel modulator by minimizing a multi-label classification loss. Let $L_{cls}(x)$ be the loss of the modulator when its input is $x$. We train our modulator by minimizing
\begin{equation}
  \begin{aligned}
  L_m = \mathbb{E}_{I_i \sim \mathcal{I}} \left [L_{cls}(F^t)\right].
  \end{aligned}
  \label{eqn:train modulator}
\end{equation}

After optimizing Eqn.~\eqref{eqn:train modulator}, the parallel modulator will be able to recognize the objects with remaining regions while being unable to recognize those whose regions are completely mined. 

In the second stage, we train the category-aware generator so that it can continue or stop mining regions for each object according to whether there are remaining regions. Concretely, the category-aware generator takes in $F^t$ and outputs the region map $\tilde{M}^t_j$ for category $j$:
\begin{equation}
  \begin{aligned}
  \tilde{M}^t_j = 1 - \frac{H^t_j- \min\{H^t_j\}}{\max\{H^t_j\}-\min\{H^t_j\}+\epsilon}.
  \end{aligned}
  \label{eqn:mask generation}
\end{equation}

Here $H^t_j$ is output from the $j$-th channel of the last convolutional layer of the generator, the division is an element-wise operation, the maximization and minimization are conducted across all spatial locations of $H^t_j$, and $\epsilon$ takes a small value and is introduced for computational stability. In Eqn.~\eqref{eqn:mask generation}, we use normalization rather than Sigmoid activation because the latter often causes training difficulty. {We then further mask $F^t$ by $\tilde{M}^t_j$, obtaining $\tilde{F}^t$:}
\begin{equation}
  \begin{aligned}
  \tilde{F}^t=F^t \odot \min\limits_ {j}\{\tilde{M}^t_j\}.
  \end{aligned}
\end{equation} 

Then we maximize the classification loss of the parallel modulator with $\tilde{F}^t$ as its input. But there may be a trivial solution for the generator, \ie to generate a region map masking the whole feature map. Such a region map actually mines everything from the input image. To avoid this issue, we add an additional regularization over the size of the mined regions: 

\begin{equation}
\begin{aligned}
L_{reg} = -\frac{1}{|\mathcal{C}_{pos}|}\sum_{j\in\mathcal{C}_{pos}}\left \|\tilde{M}^t_j \right  \| _F, 
\end{aligned}
\end{equation}
where $\| \cdot \|_F$ denotes the F-norm and $|\mathcal{C}_{pos}|$ counts the elements in $\mathcal{C}_{pos}$. Thus, the overall loss of training the category-aware generator is 
\begin{equation}
  \begin{aligned}
  L_g = \mathbb{E}_{I_i \sim \mathcal{I}} \left [-L_{cls}(\tilde{F^t})+\lambda L_{reg} \right],
  \end{aligned}
  \label{eqn:train generator}
\end{equation}
where $\lambda$ is a trade-off parameter. Since we train the category-aware generator by maximizing the loss of the parallel modulator, we say the modulator ``guides'' the generator to mine regions. 

For a given object, if its regions are not completely mined in previous steps, we will make the generator generate a region map to mine the remaining regions by minimizing Eqn.~\eqref{eqn:train generator}. If there is no object region left, \ie the parallel modulator cannot recognize this object, the second term in Eqn.~\eqref{eqn:train generator} dominates in the optimization process as the first term cannot guide the generator to update. Decreasing the second term, \ie increasing the F-norm of generated region maps, pushes the category-aware generator to generate a region map whose all pixel values are equal to 1. Such a region map does not contain any object region, which indicates the region mining sub-process  of this object should stop.

In the last stage, we use the updated category-aware generator to generate region maps $M^t_j$ for each semantic category $j$ in image $I$, and store them in their corresponding region map pool $P_j$. Note that here $M^t_j$ differs from $\tilde{M}^t_j$ in that $M^t_j$ is generated by the updated generator, while $\tilde{M}^t_j$ is produced to train {the generator.} 

\begin{algorithm}[t]
  \caption{Region Mining procedure of Multi-Miner.}
  \label{alg:oarm}
  \begin{algorithmic}[1]
  \REQUIRE Training set $\mathcal{I}$, factor $\lambda$, training epochs $n_m$ and $n_g$, learning rate $\eta$.
  \STATE Initialize $\theta_e$ and $\theta_m$, then keep $\theta_e$ fixed.
  \WHILE{not all object regions are mined}
    \STATE Freeze $\theta_g$.
    \FOR{$n_m$ epochs}
      \STATE Update $\theta_m$: $\theta_m \leftarrow L_m - \eta\nabla_{\theta_m} L_m$.
    \ENDFOR
    \STATE Freeze $\theta_m$.
    \FOR{$n_g$ epochs}
      \STATE Update the $\theta_g$ under the guidance the modulator: $\theta_g \leftarrow L_g - \eta\nabla_{\theta_g} L_g$.
    \ENDFOR
    \FOR{$I_i$ in $\mathcal{I}$}
     \IF{{$M^t_j$ contains object regions}}
        \STATE Store $M^t_j$ in $P_j$.
      \ENDIF
    \ENDFOR
  \ENDWHILE
  \ENSURE  Region map pools $P_j$ for all images $I_i$.
  \end{algorithmic}
  \end{algorithm}
\paragraph{Training process}
We first minimize the classification loss w.r.t. $\theta_e$ and $\theta_m$ over $\mathcal{I}$ to obtain the initial feature extractor and parallel modulator. Then, we only update $\theta_m$ and $\theta_g$, keeping $\theta_e$ fixed for computational efficiency. The whole region mining procedure is summarized in Alg.~\ref{alg:oarm}. After all the region mining sub-processes stop, we merge all the region maps {for each object} and obtain the mined regions $M^f_j$ by
\begin{equation}
  \begin{aligned}
  M^f_j = \min\limits_{\tau} \{ M^\tau_j\},
  \end{aligned}
\end{equation}
where $\tau = 1,\cdots, T_j$, and $T_j$ is the last step of category $j$.

\subsection{Multi-scale training strategy}
Due to the decreased spatial resolution with stacked convolutional layers, some background regions, particularly those surrounding the object regions, would be inevitably included during the region mining procedure. To mitigate this problem, we propose a multi-scale training strategy to progressively {mine} finer object regions.

Specifically, we start the region mining procedure with input images of a small scale, \ie low resolution. This helps our multi-miner to obtain holistic information of the object regions, and mine the main part of the objects. Then, we increase the spatial resolution of the input images step by step, so that our multi-miner can progressively mine details of the object regions. In this way, the region maps become increasingly finer, and backgrounds are kept from being included. However, considering computational efficiency, the spatial resolution cannot be increased without any constraint. Thus we only use a set of $K$ different spatial scales. From the first region mining step to the $K$-th step, we gradually increase the spatial resolution of the input images. From the $K$-th step, the spatial resolution stops changing.

Formally, let $\mathcal{S} = \{s_1,\cdots,s_K\ | s_1 < \cdots < s_K ,K\geqslant2\}$ denote the set of spatial resolution. The spatial resolution of the input images in the $t$-th step $r_t$ is given by
\begin{equation}
r_t=\left\{
\begin{aligned}
&s_t, & \text{if}\ \ t \leqslant K;\\
&s_K, & \text{if}\ \ t > K.
\end{aligned}
\right.
\end{equation}

Note that due to multi-scale training, in Eqn.~\eqref{eqn:feature masking}, some of the previously generated region maps are of lower spatial resolution than the feature map $F$. We use bilinear interpolation to upsample these region maps so that they have the same spatial resolution as $F$.

\subsection{Connection with distribution mapping} 
In this subsection, we reveal the connection between region mining and distribution mapping. With such a connection, we can explain why our multi-miner is able to perform region mining that is adaptive to each single object.

A multi-category region mining task can be recast as multiple binary region mining tasks of two categories, \ie foreground and background.
Without loss of generality, we consider a binary region mining task. The goal of such a task is to mine all foreground regions from the input image, and leave only background regions in the image. Namely, binary region mining aims to map the distribution of the images containing both foregrounds and backgrounds to the distribution of the images containing only backgrounds.

Formally, we define two original distributions: $p_0$ is the distribution of the images containing only backgrounds, and $p_1$ is the distribution of the images that contain both foregrounds and backgrounds. We denote an image as $x$.
Then, region mining aims to find a mapping $\mathcal{M}(\cdot)$ \st $\forall x \sim p_1, \mathcal{M}(x) \sim p_0$. 
As proven in \cite{gan}, this can be achieved by utilizing a generator $G$ to implement the mapping $\mathcal{M}(\cdot)$ and a discriminator $D$ to differentiate samples from $p_0$ and samples generated by $G$. We further denote the distribution of the images produced by the generator $G$ as $q_1$ , \ie $G(x) \sim q_1$. Similar to \cite{gan}, it can be proven that by solving the following minimax objective optimization problem:
\begin{equation}
\begin{aligned}
\min_G \max_D \mathbb{E}_{x \sim p_0} \log[1-D(x)] + \mathbb{E}_{x \sim p_1} \log[D(G(x))],
\end{aligned}
\label{eqn:gan loss}
\end{equation}
we reach an equilibrium where both G and D stop updating, and have $q_1 = p_0$, meaning we obtain a generator $G$ which maps all images containing both foregrounds and backgrounds to the images containing only backgrounds.

For a $|\mathcal{C}|$-category region mining task, however, we do not need to instantiate $|\mathcal{C}|$ pairs of  generator and discriminator for all categories. Instead, we can leverage two neural networks to model the $|\mathcal{C}|$ generators and the $|\mathcal{C}|$ discriminators respectively, which correspond to the category-aware generator and the parallel modulator in our multi-miner. 
Namely, our multi-miner is actually the combination of the $|\mathcal{C}|$ pairs of generator and discriminator for all categories. With such a mechanism, the multi-miner performs region mining sub-processes for each category in each image, and thus is adaptive to each single object.


\section{Experiment}
\subsection{Settings}
\paragraph{Dataset and evaluation metrics}
 We evaluate our multi-miner for weakly-supervised semantic segmentation on the PASCAL VOC 2012  benchmark~\cite{voc}. It provides  images from 20 object categories and is split into training (1,464 images), validation (1,449 images) and testing (1,456 images). We use the augmented training set provided by \cite{vocaug} that contains 10,582 images to train our multi-miner based region mining model as well as the segmentation model. We compare our model with state-of-the-arts on both the validation and test sets. 
 The segmentation performance is evaluated in terms of pixel Intersection-over-Union (IoU) averaged on 21 semantic categories. The results on the test set are obtained by submitting the predicted results to the official PASCAL VOC evaluation server.
 
\paragraph{Implementation}
Following previous works~\cite{ae, seenet, gain}, we use VGG-16~\cite{vgg} pre-trained on ImageNet~\cite{imagenet} to build our multi-miner.
Specifically, for its feature extractor, we remove the layers after \textit{conv5-3}. 
Furthermore, we modify \textit{pool1}, \textit{pool2} and \textit{pool3} by changing  their kernel size to $3$, stride to $2$, and padding to $1$. Similarly, we modify \textit{pool4} and \textit{pool5} so that their kernel size is $3$, stride is $1$, and padding is $1$. With such modifications to the aforementioned pooling layers, given the same input image, the spatial resolution of the feature maps output by \textit{conv5} is $2\times $ larger than that of the \textit{conv5} feature maps from the vanilla VGG-16. 
The parallel modulator of our multi-miner consists of three convolution layers, among which the first two have 1024 $3\times3$ convolution kernels and the third one has 20 $1\times1$ kernels, corresponding to 20 object categories. After the convolutional layers, there is a global average pooling layer to output the classification scores.
The category-aware generator has the same convolutional layers as the parallel modulator and has a ReLU activation layer.

We initialize the feature extractor and parallel modulator with input size of $321^2$, batch size of 64, and weight decay of 0.0001. The initial learning rate is 0.001 for the feature extractor layers and 0.01 for the modulator layers, both divided by 10 after 30 epochs. We train this classification network for totally 50 epochs. The parameters of the feature extractor are frozen afterwards.
Then we initialize our category-aware generator using the modulator parameters.

For computation efficiency, before the region mining procedure, we first extract and store the features output by the feature extractor, and then use the features to train the parallel modulator and the category-aware generator. Specifically, we train the modulator and the generator for 15 epochs and 1 epoch respectively, with weight decay of $10^{-4}$ and learning rate of $10^{-3}$. As for multi-scale training, the spatial resolutions are $256^2$, $321^2$ and $417^2$ for the first, second and third step respectively, and we keep $417^2$ from the third step. The corresponding batch size for the three scales are 256, 128 and 64.

For the segmentation task, to fairly compare with other works, we adopt the standard Deeplab-LargeFOV architecture~\cite{deeplabv1} pretrained on ImageNet~\cite{imagenet}. Following~\cite{seenet, dcsp, dsrg, mcof, ficklenet}, we also use ResNet \cite{resnet} version of Deeplab-LargeFOV architecture~\cite{deeplabv2} and report the results of both versions. We use the same background cues as SeeNet~\cite{seenet}. When using Conditional Random Fields (CRF) for post-processing, we adopt the same code as in~\cite{deeplabv1}.

\begin{table}[!t]
\centering
 \setlength{\tabcolsep}{1.8mm}{
\begin{tabular}{lccc}
\toprule
Methods & Training Data & \textit{val} & \textit{test} \\
\midrule
\multicolumn{4}{l}{\emph{Backbone: VGG-16}} \\
EM-Adapt \scriptsize{ICCV '15} \cite{pthk1} & 10K & 38.2 & 39.6 \\
DCSM \scriptsize{ECCV '16} \cite{dcsm} & 10K & 44.1  & 45.1 \\
SEC \scriptsize{ECCV '16} \cite{sec} & 10K & 50.7  & 51.7\\
Oh \etal \scriptsize{CVPR '17}\cite{oh}  & 10K & 55.7  & 56.7 \\
AE-PSL \scriptsize{CVPR '17} \cite{ae}& 10K & 55.0  & 55.7 \\
TPL \scriptsize{ICCV '17} \cite{tpl} & 10K & 53.1  & 53.8 \\
DCSP \scriptsize{BMVC '17} \cite{dcsp} & 10K & 58.6  & 59.2  \\
GAIN \scriptsize{CVPR '18} \cite{gain} & 10K & 55.3  & 56.8 \\
MDC \scriptsize{CVPR '18} \cite{mdc} & 10K & 60.4  & 60.8  \\
DSRG \scriptsize{CVPR '18} \cite{dsrg} & 10K & 59.0  & 60.4 \\
MCOF \scriptsize{CVPR '18} \cite{mcof} & 10K & 56.2  & 57.6 \\ 
SeeNet \scriptsize{NeuralIPS '18} \cite{seenet} & 10K & 61.1  & 60.7 \\
FickleNet \scriptsize{CVPR '19}  \cite{ficklenet} & 10K & 61.2  & 61.9 \\
SSNet \scriptsize{ICCV '19}  \cite{jointsal} & 10K & 57.1  & 58.6  \\
Ours & 10K & \textbf{62.8} & \textbf{63.2}\\
\bottomrule
\multicolumn{4}{l}{\emph{Backbone: ResNet-101}} \\
DCSP \scriptsize{BMVC '17} \cite{dcsp} & 10K & 60.8 & 61.9 \\
DSRG \scriptsize{CVPR '18} \cite{dsrg} & 10K & 61.4 & 63.2 \\
MCOF \scriptsize{CVPR '18} \cite{mcof} & 10K & 60.3 & 61.2 \\ 
SeeNet \scriptsize{NeuralIPS '18} \cite{seenet} & 10K & 63.1  & 62.8 \\
FickleNet \scriptsize{CVPR '19} \cite{ficklenet} & 10K & 64.9  & 65.3 \\
Ours & 10K & \textbf{65.9}  & \textbf{66.1} \\
\bottomrule
\end{tabular}}
\caption{Comparison of weakly-supervised semantic segmentation methods on VOC 2012 validation and test set.}
\vspace{-2mm}
\label{tbl:vocresults}
\end{table}

\subsection{Comparisons with state-of-the-arts}
We conduct experiments to compare our proposed model with existing state-of-the-art weakly-supervised semantic segmentation ones under the same setting. Table~\ref{tbl:vocresults} shows the experiment results.

We observe that our multi-miner outperforms all the baselines. Among all the baseline methods, the erasing-based methods, \ie AE~\cite{ae}, GAIN~\cite{gain} and SeeNet~\cite{seenet}, share some similarities with ours. The difference is that they mine regions of all objects with a constant number of erasing steps, while our proposed multi-miner is object-adaptive, taking different numbers of region mining steps for different objects. This property helps the multi-miner to have performance gains of 7.8\%, 7.5\%, 1.7\% on \textit{val} set with the standard Deeplab-LargeFOV architecture over AE, GAIN and SeeNet respectively. FickleNet~\cite{ficklenet} is not object-adaptive either, and needs to process each image for a large number of times to obtain the regions. 

\begin{figure*}[!t]
\centering 
\includegraphics[width=\textwidth]{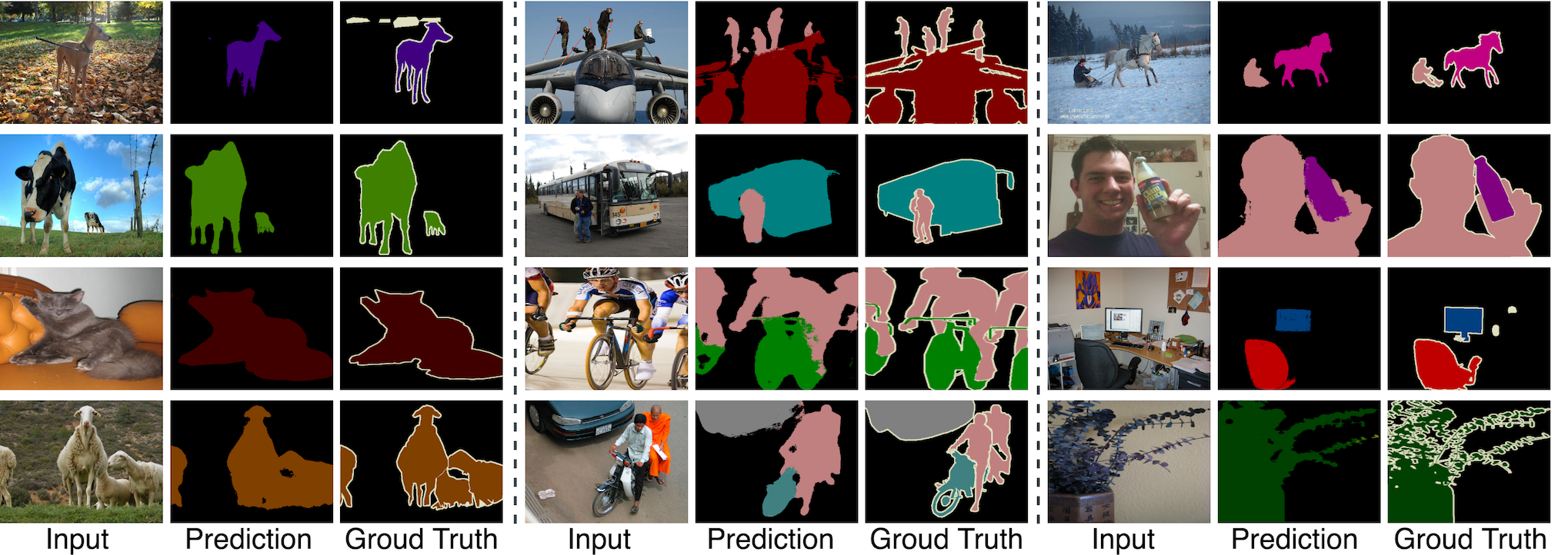}
\vspace{-3mm}
\caption{Qualitative segmentation results on the PASCAL VOC 2012 validation set. Please refer to Supplementary Material for more results.} 
\label{fig:segresults} 
\end{figure*}

\begin{figure}[!t]
\centering 
\begin{overpic}[width=0.48\textwidth]{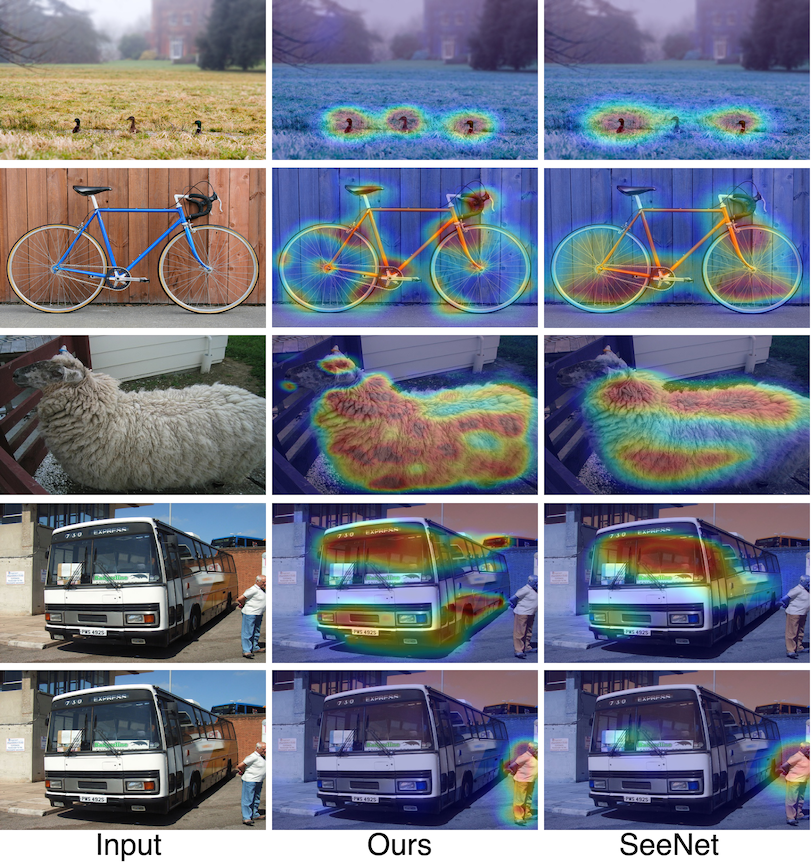}
\put(0.5, 19.5){\addTexB{\sffamily person}}
    \put(0.5, 38.5){\addTexB{\sffamily bus}}
    \put(0.5, 58){\addTexB{\sffamily sheep}}
    \put(0.5, 77){\addTexB{\sffamily bicycle}}
    \put(0.5, 96.5){\addTexB{\sffamily bird}}
\end{overpic}
\caption{Qualitative comparisons of region mining results between ours and SeeNet. Please refer to Supplementary Material for more results.} 
\label{fig:seenet comparisons} 
\vspace{-3mm}
\end{figure}
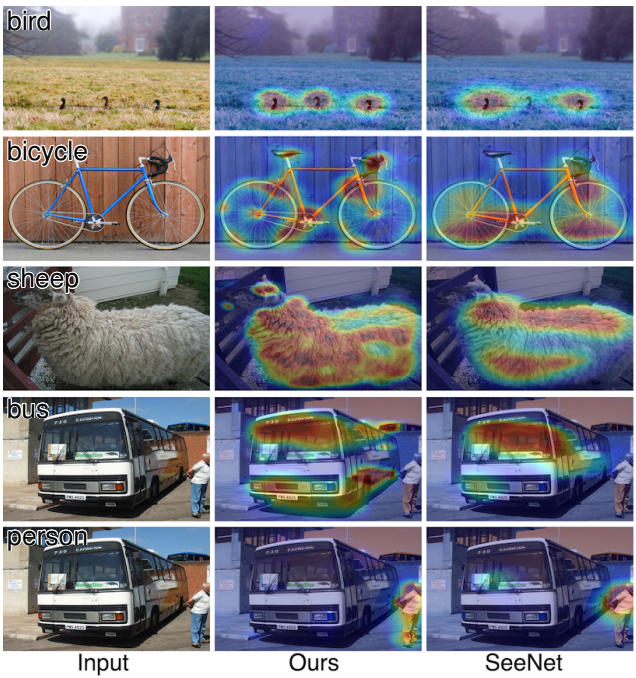

Some qualitative segmentation results are shown in Fig.~\ref{fig:segresults}. We can see that the segmentation network performs satisfyingly with the supervision of pseudo masks produced by our multi-miner, and can produce complete and accurate areas even for complex images. Moreover, in Fig.~\ref{fig:seenet comparisons}, we qualitatively compare the regions mined by our proposed multi-miner and those by SeeNet~\cite{seenet}. It can be observed that due to the object-adaptability of our method, the mined regions of our method are more integral and less contaminated by background regions.

\subsection{Ablation studies}
\paragraph{Comparisons across steps}
In our experiment, we find that the longest region mining sub-process takes 7 steps. 

After all region mining sub-processes stop, we merge the region maps generated from the first $T$ steps $(T = 1,\cdots,7)$ for each object, and train the standard Deeplab-LargeFOV architecture accordingly. Concretely, for the category $j$ in image $I$, we merge $\{M_j^\tau\}_{\tau=1}^{T_j}$ if $T_j < T$, and merge $\{M_j^\tau\}_{\tau=1}^{T}$ otherwise, where $T_j$ is last step of category $j$.

\begin{figure}[!t]
\centering 
\includegraphics[width=0.48\textwidth]{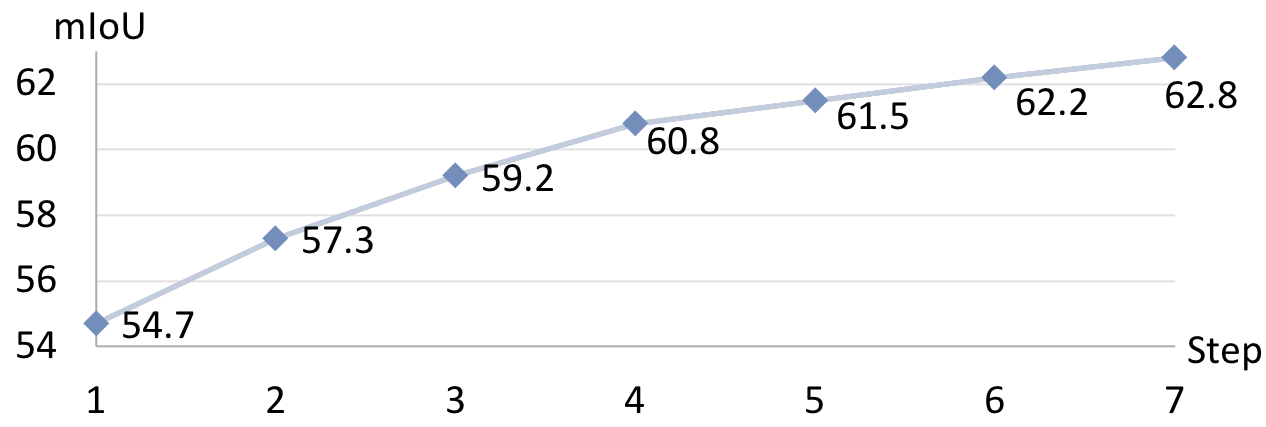}
\caption{Relationship between segmentation mIoU scores on the PASCAL VOC 2012 \textit{val} set and the number of steps for merging.} 
\label{fig:iou steps} 
\end{figure}

We summarize their performance on the PASCAL VOC 2012 validation set in Fig.~\ref{fig:iou steps}. We can see that the performance increases as $T$ increases from 1 to 7. This demonstrates our multi-miner gradually mines object regions from the 1st to the 7th step. In addition, we can see that the increment of mIoU per step begins to decrease after the 4th step. The reason may be that a majority of objects are roughly or completely mined before the 4th step.

\begin{figure*}[!t]
\centering 
\begin{overpic}[width=\textwidth]{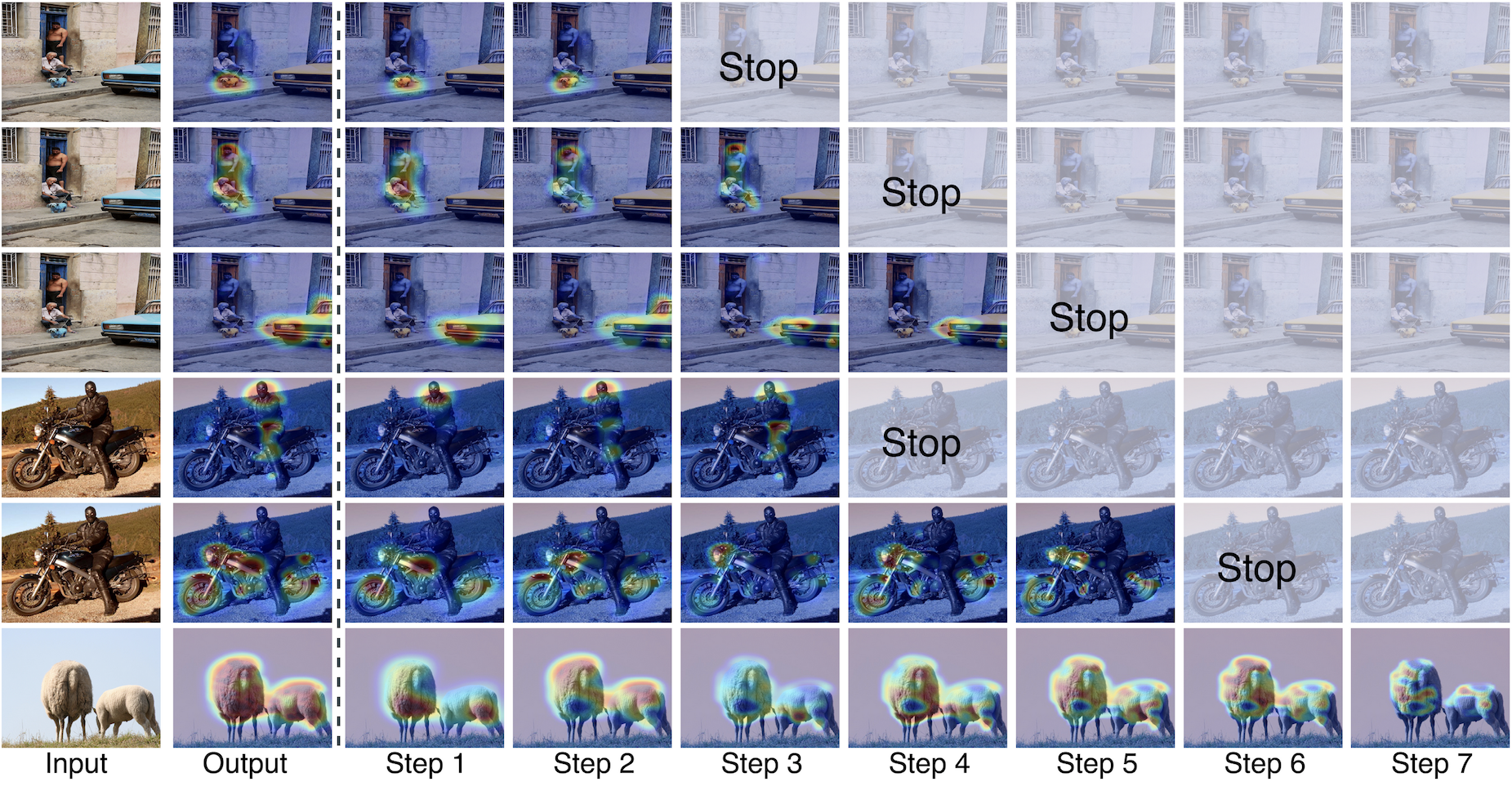}
    \put(0.3, 9){\addTexB{\footnotesize \sffamily sheep}}
    \put(0.3, 17.5){\addTexB{\footnotesize \sffamily motor}}
    \put(0.3, 26){\addTexB{\footnotesize \sffamily person}}
    \put(0.3, 34.2){\addTexB{\footnotesize \sffamily car}}
    \put(0.3, 42.5){\addTexB{\footnotesize \sffamily person}}
    \put(0.3, 50.5){\addTexB{\footnotesize \sffamily dog}}
\end{overpic}
\caption{Regions mined in different steps. In each row, the first column is the input image, the second column is the output region mining result, and the following rows show the mined region in every step. Here ``Stop'' means the region mining sub-process of this object stops, and the semi-transparent region maps means there is no more remaining object region. Note that the number of region mining steps of different objects in the same image also varies. Please refer to Supplementary Material for more results.} 
\label{fig:steps} 
\end{figure*}

Moreover, we visualize some of the region mining sub-processes in Fig.~\ref{fig:steps}. We observe that our multi-miner adaptively takes different region mining steps for different objects. Generally, larger objects need more steps. The first two rows and the 3rd-5th rows show two examples of images containing multiple objects. From them, we can observe that the number of region mining steps for different objects in the same image also differs. 

 
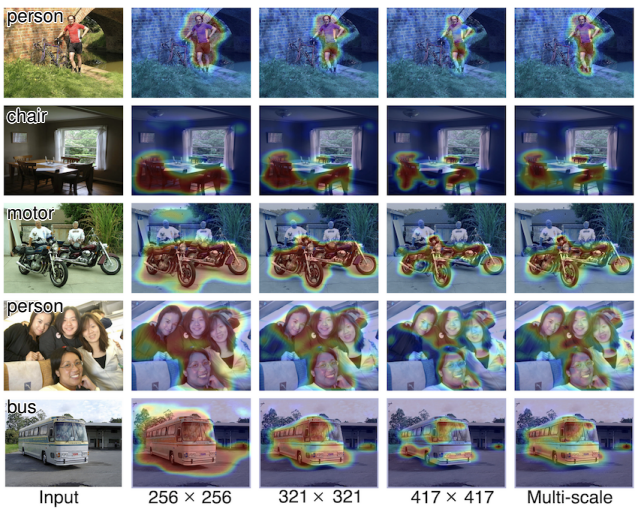
\begin{figure}[!t]
\centering 
\begin{overpic}[width=0.48\textwidth]{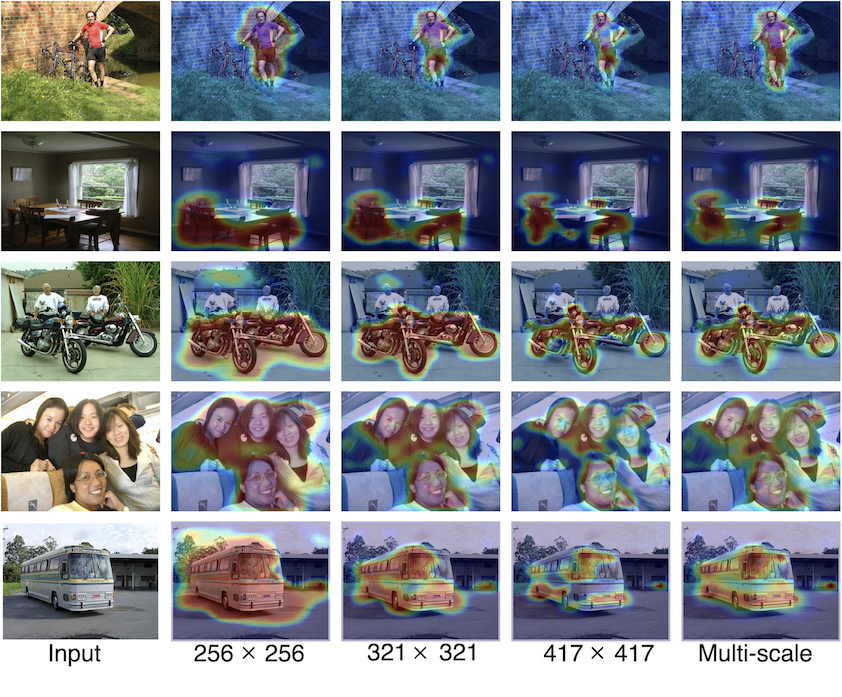}
    \put(0.5, 18.5){\addTexB{\scriptsize \sffamily bus}}
    \put(0.5, 34.5){\addTexB{\scriptsize \sffamily person}}
    \put(0.5, 49.2){\addTexB{\scriptsize \sffamily motor}}
    \put(0.5, 64.6){\addTexB{\scriptsize \sffamily chair}}
    \put(0.5, 80.5){\addTexB{\scriptsize \sffamily person}}
\end{overpic}
\caption{Region mining results of different training strategies. For single-scale baselines, the regions become finer but smaller as the scale of input images increases. 
Multi-scale training gives good results for both large and small objects. Better viewed in color.} 
\label{fig:multi comparison} 
\end{figure}

\paragraph{Effectiveness of multi-scale training} To prove effectiveness of our multi-scale training strategy, we compare it with 3 single-scale baselines. Specifically, we train 3 multi-miner models with input images of spatial resolution fixed at $256^2$, $321^2$, $417^2$ respectively, and compare their results with that of the multi-miner trained with multi-scale strategy. Fig.~\ref{fig:multi comparison} shows some of the mined regions. Comparing the three baselines, we can observe with a fixed small scale \ie $256^2$, our multi-miner can mine a large size of object regions, but the regions are coarse and include some irrelevant background regions, particularly for small objects. With a fixed large scale \ie $417^2$, the mined regions are finer but less integral for large objects. In contrast, our multi-scale training strategy inherits the advantages of both small-scale and large-scale training, but does not have their drawbacks, and thus is good for both large and small objects and gives integral object regions with fine boundaries.

Additionally, we train the standard Deeplab-LargeFOV architecture with object regions obtained with the above four training strategies. The segmentation results on the PASCAL VOC 2012 validation set are summarized in Table~\ref{tbl:results scales}. It can be seen that our multi-miner achieves the best performance with multi-scale training, which further demonstrates the advantage of multi-scale training.

\begin{table}[t]
 \centering
\begin{tabular}{l|cccc}
 \toprule 
 Training Scale & $256^2$ & $321^2$ &$417^2$ & Multi-scale \\
 \midrule
  mIoU(\%) & 59.8 & 62.1 & 61.7 & 62.8 \\
 \bottomrule
 \end{tabular}               
  \caption{Comparison of segmentation mIoU scores using regions mined by the multi-miner with different training strategies. }
 \label{tbl:results scales}
 \end{table}

\section{Conclusion}
We proposed an object-adaptive multi-miner framework to mine integral and fine object regions. It adaptively takes different numbers of region mining steps for different objects, offering high-quality region mining results. Moreover, we proposed a multi-scale training strategy to mine progressively finer regions. We applied the mined regions to training a weakly-supervised semantic segmentation model. Experiment results have shown that our proposed model offers much higher-quality mined object regions than state-of-the-arts.

{\small
\bibliographystyle{ieee_fullname}
\bibliography{egbib}
}

\end{document}